\documentclass[11pt,twoside]{article}
\usepackage[utf8]{inputenc}
\usepackage{graphicx}
\usepackage{amsmath, amssymb, bm}
\usepackage{hyperref}
\usepackage{natbib}
\usepackage{fancyhdr}
\usepackage{geometry}
\usepackage{titlesec}
\usepackage{lipsum}
\usepackage{orcidlink}
\usepackage{booktabs}
\usepackage{placeins}
\usepackage{float}
\usepackage{array}
\usepackage[ruled,vlined,linesnumbered]{algorithm2e}
\usepackage{amsmath,amssymb,amsfonts,bm,mathtools}

\setcitestyle{authoryear}
\newcommand{\R}{\mathbb{R}}
\newcommand{\E}{\mathbb{E}}
\newcommand{\KL}{\mathrm{KL}}

\newcommand{\lse}{\mathrm{LSE}}

\hypersetup{
colorlinks=true,
linkcolor=blue,
filecolor=blue,
citecolor=blue,
urlcolor=black
}

\fancypagestyle{plain}{
\fancyhf{}

}

\title{
\vspace{-1.5em}
\hrule height 1.5pt
\vspace{0.8em}
SAHMM-VAE: A Source-Wise Adaptive Hidden Markov Prior Variational Autoencoder for Unsupervised Blind Source Separation
\vspace{0.8em}
\hrule height 1.5pt
\vspace{1em}
}

\author{%
\begin{minipage}[t]{.48\textwidth}\centering\small
  \textbf{Yuan-Hao Wei\orcidlink{0000-0001-9439-0780}}\\
  \texttt{Yuan-Hao.Wei@outlook.com; yuan-hao.wei@connect.polyu.hk}
\end{minipage}
}

\date{}

\begin{document}
\maketitle
\thispagestyle{plain}

\begin{abstract}
We propose SAHMM-VAE, a source-wise adaptive Hidden Markov prior variational autoencoder for unsupervised blind source separation. Instead of treating the latent prior as a single generic regularizer, the proposed framework assigns each latent dimension its own adaptive regime-switching prior, so that different latent dimensions are pulled toward different source-specific temporal organizations during training. Under this formulation, source separation is not implemented as an external post-processing step; it is embedded directly into variational learning itself. The encoder, decoder, posterior parameters, and source-wise prior parameters are optimized jointly, where the encoder progressively learns an inference map that behaves like an approximate inverse of the mixing transformation, while the decoder plays the role of the generative mixing model. Through this coupled optimization, the gradual alignment between posterior source trajectories and heterogeneous HMM priors becomes the mechanism through which different latent dimensions separate into different source components. To instantiate this idea, we develop three branches within one common framework: a Gaussian-emission HMM prior, a Markov-switching autoregressive HMM prior, and an HMM state-flow prior with state-wise autoregressive flow transformations. Experiments show that the proposed framework achieves unsupervised source recovery while also learning meaningful source-wise switching structures. More broadly, the method extends our structured-prior VAE line from smooth, mixture-based, and flow-based latent priors to adaptive switching priors, and provides a useful basis for future work on interpretable and potentially identifiable latent source modeling.

\end{abstract}


\section{Introduction}

Many real source signals are not governed by a single stationary law. They alternate between activity patterns, switch between local regimes, and exhibit temporal dependence that changes from segment to segment. From the viewpoint of blind source separation (BSS), this observation is important because such switching structure is often not a nuisance but part of the information that distinguishes one source from another. If a latent model ignores this structure and forces all hidden components to share one simple prior, then it may still reconstruct the observations well, yet remain poorly matched to the heterogeneous source processes that make separation possible in the first place.

BSS seeks to recover latent source signals from observed mixtures without direct access to the mixing mechanism. Independent component analysis (ICA) established one of the most influential solutions by showing that, under suitable linear assumptions, statistical independence and non-Gaussianity can be sufficient for source recovery \citep{comon1994independent,hyvarinen1997fast,hyvarinen2000independent}. Once the setting becomes nonlinear, noisy, or temporally structured, however, the problem becomes much harder and the role of latent structural assumptions becomes substantially more central.

Variational autoencoders (VAEs) \citep{kingma2014auto} provide a natural probabilistic language for revisiting this problem. In a source-separation interpretation, the encoder can be read as a demixing map from mixtures to latent components, while the decoder acts as a generative remixing map from latent components back to observations \citep{wei2024innovative}. This perspective has become increasingly meaningful in light of identifiable nonlinear ICA and identifiable VAE formulations, which show that latent recovery is deeply tied to the structure imposed on latent variables and their conditional laws \citep{hyvarinen2017nonlinear,hyvarinen2019nonlinear,khemakhem2020variational}. In other words, deep encoder--decoder flexibility alone is not enough; the latent prior matters because it determines what kinds of hidden variables the model prefers to infer.

This is precisely where standard VAE practice becomes limiting for BSS. A shared isotropic Gaussian prior is convenient, but it is also deliberately featureless. It does not distinguish between a source that is smooth, one that is multimodal, one that exhibits temporal persistence, and one that repeatedly switches among local dynamic modes. For source separation, such symmetry can be counterproductive: if all latent dimensions are encouraged toward the same simple prior, then the model is given little incentive to let different dimensions specialize into genuinely different source processes.

Our work has therefore treated prior design not as a background regularization choice but as the main vehicle for source-oriented latent modeling. In PAVAE, adaptive Gaussian-process priors were used to encode source-wise smoothness and temporal correlation \citep{wei2024innovative}. In PDGMM-VAE, each latent dimension was assigned its own adaptive Gaussian mixture prior to capturing different source dimensions that were not Gaussian \citep{wei2026pdgmmvae}. AR-Flow VAE further expanded this line by equipping each latent source with a parameter-adaptive autoregressive flow prior to model richer ordered dependence and non-Gaussian temporal structure \citep{wei2026arflow}. A consistent lesson emerging from these studies is that, when each latent dimension is treated explicitly as a source candidate, heterogeneous prior constraints can drive different latent dimensions toward different source roles.

The present paper extends that program from continuously varying source structure to switching source structure. We propose \emph{SAHMM-VAE}, a \emph{source-wise adaptive Hidden Markov prior variational autoencoder} for unsupervised blind source separation. The key modeling move is simple but consequential: rather than assigning all latent variables one common prior, we equip each latent dimension with its own adaptive hidden-state process. Each source dimension therefore carries its own regime-switching bias, and different latent dimensions are encouraged to settle into different switching organizations. This is especially natural when sources are better described by alternating regimes than by a single static density.

To make this idea operational, we develop a unified family with three representative branches. The first uses a Gaussian-emission HMM prior and captures source-wise switching among state-dependent marginal distributions. The second upgrades each source prior to a Markov-switching autoregressive process, thereby incorporating state-dependent linear temporal dynamics. The third further enriches the state-conditional dynamics by using state-wise invertible flow transformations, allowing non-Gaussian innovations within each regime. These are not three unrelated models, but three increasingly expressive realizations of the same source-wise switching-prior principle.

This work should also be distinguished from neighboring lines of research. It is not classical HMM-based source separation in the factorial-HMM tradition, which typically performs source inference outside a VAE formulation \citep{ghahramani1997factorial,reyes2003multichannel}. It is also different from Hidden Markov nonlinear ICA, whose main focus is unsupervised nonlinear ICA and identifiability through a latent Markov chain rather than source-wise VAE prior design \citep{halva2020hidden}. Nor is it the same as speech-oriented HMM-VAE models for acoustic unit discovery, where HMMs organize discrete speech units rather than act as source-specific priors for latent source recovery \citep{ebbers2017hidden,glarner2018full}. Our emphasis is instead on using source-wise adaptive switching priors to make latent dimensions separate as source processes within one unified encoder--decoder training objective.

A further point deserves emphasis. In this paper, \emph{unsupervised} does not merely mean that ground-truth sources are absent from the labels. The separation process itself is embedded in training. As the reconstruction term and the prior-matching term are optimized jointly, the posterior trajectories, decoder parameters, and source-wise prior parameters co-adapt. Different latent dimensions gradually become associated with different prior parameters; those priors, in turn, pull the posterior trajectories toward different structured source explanations through the KL term. Thus, optimization of the loss function is simultaneously the process by which the sources become separated.

The main contributions of this paper are threefold. First, we introduce a source-oriented VAE framework in which each latent dimension is assigned its own adaptive HMM prior. Second, we unify three increasingly expressive switching-prior branches within one training structure, ranging from Gaussian-emission HMMs to state-flow HMM priors. Third, we show experimentally that this source-wise switching-prior design supports accurate unsupervised source recovery while also learning interpretable hidden temporal structure, thereby broadening the methodological scope of structured-prior VAE research for BSS.

\section{Related Work}

Research on blind source separation has long been shaped by ICA and its extensions. Classical ICA exploits independence and non-Gaussianity to recover latent components from mixtures and remains one of the central theoretical viewpoints in source separation \citep{comon1994independent,hyvarinen1997fast,hyvarinen2000independent}. In nonlinear settings, however, identifiability is no longer automatic, and substantial work has focused on what additional structure is needed to make nonlinear source recovery possible \citep{hyvarinen2017nonlinear,hyvarinen2019nonlinear}. This broader literature is directly relevant here because it clarifies that latent recovery depends not only on the observation model but also on structural assumptions imposed on hidden variables.

A particularly important bridge between nonlinear ICA and deep latent-variable models was established by Khemakhem et al., who showed that identifiable nonlinear ICA can be expressed within a VAE-style probabilistic framework when suitable factorization and conditioning assumptions are imposed on the latent prior \citep{khemakhem2020variational}. From the viewpoint of BSS, this result is conceptually significant: it suggests that the prior is not just a mathematical convenience for variational training, but one of the main mechanisms through which source recovery may become structured, interpretable, and potentially identifiable.

Temporal and switching structure have also played a central role in recent nonlinear ICA research. Temporal dependence and nonstationarity have been used repeatedly as signals that help break the ambiguities of nonlinear source models \citep{hyvarinen2017nonlinear,hyvarinen2019nonlinear}. Hidden Markov Nonlinear ICA is especially relevant, because it replaces observed segment indices by an inferred hidden-state sequence and thereby achieves fully unsupervised nonlinear ICA under a Markovian latent-state process \citep{halva2020hidden}. That work is highly relevant in spirit, but its primary objective differs from ours. It studies how a latent Markov chain can enable unsupervised nonlinear ICA and identifiability, whereas the present paper studies how \emph{each source dimension} in a VAE can be assigned its own adaptive HMM prior so that different latent dimensions are biased toward different switching source processes.

There is also prior work directly combining HMMs and VAEs. Ebbers et al. proposed an HMM-VAE for acoustic unit discovery, and Glarner et al. further developed a Bayesian HMM-VAE with a Dirichlet-process prior over acoustic units \citep{ebbers2017hidden,glarner2018full}. These studies show that HMM and VAE components can be integrated effectively, but the latent semantics are different. In acoustic unit discovery, the HMM is used to model discrete units in speech; in our setting, the HMM is assigned source-wise and serves as a structured prior over continuous latent source trajectories for BSS.

Another neighboring tradition is HMM-based source separation outside the VAE framework. Factorial HMMs and related models assign separate hidden-state processes to different sources and have been used for mixture modeling and source recovery \citep{ghahramani1997factorial,reyes2003multichannel}. This literature is important because it confirms that giving different sources separate Markov structure is a natural modeling principle. Nevertheless, these methods are methodologically distinct from the present study: they are not based on encoder--decoder variational inference, do not frame the problem through a VAE objective, and do not focus on source-wise adaptive prior learning inside a deep generative model.

Finally, the present paper continues our own structured-prior VAE line for BSS and nonlinear ICA. PAVAE introduced adaptive Gaussian-process priors to model source-wise smoothness and temporal correlation \citep{wei2024innovative}. PDGMM-VAE then moved toward per-dimension heterogeneous priors by assigning each latent component its own adaptive Gaussian mixture model \citep{wei2026pdgmmvae}. AR-Flow VAE further enriched this direction by equipping each latent source with a parameter-adaptive autoregressive flow prior \citep{wei2026arflow}. In a related but architecturally different branch, Half-VAE explored ICA-style latent recovery without an encoder by directly optimizing latent variables as trainable parameters \citep{wei2024halfvae}. Relative to these works, the present study fills the switching-prior gap in this broader research program: it asks what happens when each latent source is driven not by a smooth, mixture-based, or continuously transformed prior, but by its own adaptive regime-switching process.

\section{Methodology}
\label{sec:methodology}

\subsection{Source-wise switching latent formulation}
\label{subsec:switching_formulation}

We consider a sequence of observed mixtures $\mathbf{Y}=\{\mathbf{y}_t\}_{t=1}^{T}$ with $\mathbf{y}_t\in\R^{m}$. The latent representation at time index $t$ is denoted by $\mathbf{s}_t\in\R^{n}$, and the $j$th latent dimension is interpreted explicitly as the $j$th source candidate trajectory across the whole sequence. Instead of introducing a single common latent prior for all dimensions, we assign each source trajectory its own hidden-state process. The resulting latent prior is therefore decomposed \emph{source by source}, not merely sample by sample.

Under this view, the generative model is written as
\begin{equation}
p_{\theta,\psi}(\mathbf{Y},\mathbf{S})
=
p_{\theta}(\mathbf{Y}\mid\mathbf{S})\,p_{\psi}(\mathbf{S}),
\label{eq:joint_model_hmm}
\end{equation}
where $\mathbf{S}=\{\mathbf{s}_t\}_{t=1}^{T}$, $\theta$ collects decoder parameters, and $\psi$ collects the source-wise prior parameters. The decoder maps latent source samples back to the observation space through
\begin{equation}
\hat{\mathbf{y}}_t=g_{\theta}(\mathbf{s}_t),
\end{equation}
so the same formulation covers both linear and nonlinear mixing mappings depending on the choice of $g_{\theta}(\cdot)$.

The intractable posterior is approximated by a variational encoder $q_{\phi}(\mathbf{S}\mid\mathbf{Y})$. In the present implementation, the encoder produces source-wise posterior means
\begin{equation}
\bm{\mu}_t=f_{\phi}(\mathbf{y}_t)
= [\mu_{t,1},\mu_{t,2},\dots,\mu_{t,n}]^{\top},
\end{equation}
and the variational posterior is factorized as
\begin{equation}
q_{\phi}(\mathbf{S}\mid\mathbf{Y})
=
\prod_{j=1}^{n}\prod_{t=1}^{T}
\mathcal{N}(s_{t,j}\mid\mu_{t,j},\sigma_j^2),
\label{eq:qposterior_hmm}
\end{equation}
where $\sigma_j^2>0$ is a learnable variance shared across time for source dimension $j$. Sampling is performed by reparameterization:
\begin{equation}
s_{t,j}=\mu_{t,j}+\sigma_j\epsilon_{t,j},
\qquad \epsilon_{t,j}\sim\mathcal{N}(0,1).
\label{eq:reparam_hmm}
\end{equation}

This source-wise posterior parameterization is important for the present paper. The encoder does not merely produce a generic latent code. It produces candidate source trajectories, while the KL term later compares each of these trajectories with a corresponding source-specific switching prior. This posterior--prior pairing is the main mechanism by which different latent dimensions are encouraged to settle into different source roles.

The reconstruction term is implemented by
\begin{equation}
\mathcal{L}_{\mathrm{rec}}
=\sum_{t=1}^{T}\|\hat{\mathbf{y}}_t-\mathbf{y}_t\|_2^2.
\label{eq:lrec_hmm}
\end{equation}
The posterior log-density is
\begin{equation}
\log q_{\phi}(\mathbf{S}\mid\mathbf{Y})
=
\sum_{j=1}^{n}\sum_{t=1}^{T}
\left[
-\frac{1}{2}\log(2\pi)
-\frac{1}{2}\log\sigma_j^2
-\frac{(s_{t,j}-\mu_{t,j})^2}{2\sigma_j^2}
\right].
\label{eq:logq_hmm}
\end{equation}

\subsection{Source-wise adaptive HMM prior backbone}
\label{subsec:hmm_backbone}

For each source dimension $j\in\{1,\dots,n\}$, we introduce a discrete hidden-state path
\begin{equation}
\mathbf{c}_{:,j}=[c_{1,j},c_{2,j},\dots,c_{T,j}]^{\top},
\qquad c_{t,j}\in\{1,\dots,K\},
\end{equation}
with source-specific initial distribution and transition matrix,
\begin{equation}
p(c_{1,j}=k)=\pi_{j,k},
\qquad
p(c_{t,j}=b\mid c_{t-1,j}=a)=A_{j,ab}.
\end{equation}
Thus, each latent dimension carries its own Markov chain rather than sharing one global hidden-state process.

Conditioned on this hidden-state path, the source trajectory $\mathbf{s}_{:,j}$ is scored by a branch-dependent state-conditional density. In generic form,
\begin{equation}
p(\mathbf{s}_{:,j},\mathbf{c}_{:,j})
=
p(c_{1,j})
\prod_{t=2}^{T}p(c_{t,j}\mid c_{t-1,j})
\prod_{t=1}^{T}p(s_{t,j}\mid \star, c_{t,j}),
\label{eq:joint_sc_generic_new}
\end{equation}
where the symbol $\star$ indicates that the state-conditional term may or may not depend on previous source values depending on the chosen branch. Marginalizing the hidden states gives
\begin{equation}
p(\mathbf{s}_{:,j})=\sum_{\mathbf{c}_{:,j}}p(\mathbf{s}_{:,j},\mathbf{c}_{:,j}),
\end{equation}
and the full prior factorizes across source dimensions:
\begin{equation}
p_{\psi}(\mathbf{S})=\prod_{j=1}^{n}p(\mathbf{s}_{:,j}).
\label{eq:full_prior_hmm}
\end{equation}

This decomposition is the structural center of SAHMM-VAE. Different latent dimensions are separated not because a post hoc algorithm rotates them afterward, but because during training each dimension is repeatedly evaluated against its own source-specific HMM prior, whose parameters are updated jointly with the rest of the model. As the latent dimensions progressively align with different source processes, the associated prior parameters also converge to distinct configurations for different dimensions. The KL term then reinforces this source-wise differentiation by encouraging each posterior factor to stay consistent with its corresponding learned prior. Importantly, this consistency is established through a bidirectional coupling between posterior and prior during optimization: the learned priors progressively adapt to different source dynamics, while the encoder is simultaneously driven toward a representation map that behaves as an approximate inverse of the underlying mixing function \(g\), that is, an effective \(g^{-1}\)-like demixing map. In this way, source separation emerges gradually as an intrinsic consequence of unsupervised training rather than as a separate post-processing step.

\subsection{Three prior branches as a progressive family}
\label{subsec:three_branches_new}

The three branches differ only in how the state-conditioned source density is defined. This yields a clean progression from switching marginal structure to switching linear dynamics and then to switching nonlinear state-wise innovations.

\subsubsection{Branch I: Gaussian-emission HMM prior}

The first branch assumes that, once the current hidden state is known, the source value is generated from a state-specific Gaussian emission:
\begin{equation}
p(s_{t,j}\mid c_{t,j}=k)
=
\mathcal{N}(s_{t,j}\mid m_{j,k},v_{j,k}),
\qquad v_{j,k}>0.
\label{eq:b1_emit_new}
\end{equation}
The corresponding local score is
\begin{equation}
\ell^{(1)}_{t,j}(k)
=
-\frac{1}{2}
\left[
\frac{(s_{t,j}-m_{j,k})^2}{v_{j,k}}+\log(2\pi)+\log v_{j,k}
\right].
\end{equation}
Using forward recursion in log domain,
\begin{align}
\alpha^{(1)}_{1,j}(k)
&= \log\pi_{j,k}+\ell^{(1)}_{1,j}(k), \\
\alpha^{(1)}_{t,j}(k)
&= \ell^{(1)}_{t,j}(k)
+\lse_{a=1}^{K}\big(\alpha^{(1)}_{t-1,j}(a)+\log A_{j,ak}\big), \quad t\ge 2,
\end{align}
where $\lse$ denotes log-sum-exp. Then
\begin{equation}
\log p^{(1)}(\mathbf{s}_{:,j})
=
\lse_{k=1}^{K}\,\alpha^{(1)}_{T,j}(k).
\label{eq:b1_logp_new}
\end{equation}
This branch emphasizes switching amplitude statistics but does not explicitly model state-dependent autoregression.

\subsubsection{Branch II: Markov-switching autoregressive HMM prior}

The second branch keeps the source-wise Markov chain but allows each regime to carry its own linear temporal dynamics. The initial observation term is
\begin{equation}
p(s_{1,j}\mid c_{1,j}=k)
=
\mathcal{N}(s_{1,j}\mid \tilde{\mu}_{j,k},\tilde{v}_{j,k}),
\end{equation}
and for $t\ge 2$,
\begin{equation}
p(s_{t,j}\mid s_{t-1,j},c_{t,j}=k)
=
\mathcal{N}\!
\Big(
s_{t,j}\mid
\mu_{j,k}+\phi_{j,k}(s_{t-1,j}-\mu_{j,k}),
\sigma_{j,k}^2
\Big).
\label{eq:b2_trans_new}
\end{equation}
Define the state-wise autoregressive (AR) mean
\begin{equation}
\eta_{t,j,k}=\mu_{j,k}+\phi_{j,k}(s_{t-1,j}-\mu_{j,k}),
\end{equation}
so that the local score is
\begin{equation}
\ell^{(2)}_{t,j}(k)
=
-\frac{1}{2}
\left[
\frac{(s_{t,j}-\eta_{t,j,k})^2}{\sigma_{j,k}^2}+\log(2\pi)+\log\sigma_{j,k}^2
\right].
\end{equation}
The forward recursion becomes
\begin{align}
\alpha^{(2)}_{1,j}(k)
&= \log\pi_{j,k}+\ell^{(2)}_{1,j}(k), \\
\alpha^{(2)}_{t,j}(k)
&= \ell^{(2)}_{t,j}(k)
+\lse_{a=1}^{K}\big(\alpha^{(2)}_{t-1,j}(a)+\log A_{j,ak}\big), \quad t\ge 2,
\end{align}
and
\begin{equation}
\log p^{(2)}(\mathbf{s}_{:,j})
=
\lse_{k=1}^{K}\,\alpha^{(2)}_{T,j}(k).
\label{eq:b2_logp_new}
\end{equation}
Relative to Branch I, this branch can distinguish states not only by marginal level but also by state-dependent persistence.

\subsubsection{Branch III: HMM state-flow prior}

The third branch retains switching dynamics and state-wise autoregression, but replaces Gaussian innovations by state-specific invertible flow-transformed innovations. The initial term is again Gaussian,
\begin{equation}
p(s_{1,j}\mid c_{1,j}=k)
=
\mathcal{N}(s_{1,j}\mid \bar{\mu}_{j,k},\bar{v}_{j,k}),
\end{equation}
while for $t\ge2$ we define the state-conditioned AR backbone
\begin{equation}
u_{t,j,k}=\frac{s_{t,j}-a_{j,k}s_{t-1,j}}{\rho_{j,k}},
\qquad \rho_{j,k}>0.
\label{eq:u_stateflow_new}
\end{equation}
This residual is linked to standard Gaussian base noise through an invertible state-wise transformation:
\begin{equation}
u_{t,j,k}=f_{j,k}(\varepsilon_{t,j,k}),
\qquad \varepsilon_{t,j,k}\sim\mathcal{N}(0,1).
\end{equation}
Equivalently, $\varepsilon_{t,j,k}=f^{-1}_{j,k}(u_{t,j,k})$, and the conditional density is obtained by change of variables,
\begin{equation}
p(s_{t,j}\mid s_{t-1,j},c_{t,j}=k)
=
\frac{1}{\rho_{j,k}}
\mathcal{N}(f^{-1}_{j,k}(u_{t,j,k})\mid 0,1)
\left|\frac{\partial f^{-1}_{j,k}(u_{t,j,k})}{\partial u_{t,j,k}}\right|.
\label{eq:b3_density_new}
\end{equation}
In implementation, each $f_{j,k}$ is composed of one-dimensional sinh--arcsinh style layers. Writing the inverse-flow output as $\varepsilon_{t,j,k}=f^{-1}_{j,k}(u_{t,j,k})$, the local score can be expressed compactly as
\begin{equation}
\ell^{(3)}_{t,j}(k)
=
-\frac{1}{2}(\varepsilon_{t,j,k}^2+\log(2\pi))
+\log\left|\frac{\partial f^{-1}_{j,k}(u_{t,j,k})}{\partial u_{t,j,k}}\right|
-\log\rho_{j,k}.
\label{eq:b3_localscore_new}
\end{equation}
The forward recursion is then
\begin{align}
\alpha^{(3)}_{1,j}(k)
&= \log\pi_{j,k}+\ell^{(3)}_{1,j}(k), \\
\alpha^{(3)}_{t,j}(k)
&= \ell^{(3)}_{t,j}(k)
+\lse_{a=1}^{K}\big(\alpha^{(3)}_{t-1,j}(a)+\log A_{j,ak}\big), \quad t\ge 2,
\end{align}
with
\begin{equation}
\log p^{(3)}(\mathbf{s}_{:,j})
=
\lse_{k=1}^{K}\,\alpha^{(3)}_{T,j}(k).
\label{eq:b3_logp_new}
\end{equation}
This is the most expressive branch because it allows each hidden regime to induce its own non-Gaussian innovation law.

\subsection{Unified objective and the separation mechanism}
\label{subsec:objective_mechanism}

For any selected branch $b\in\{1,2,3\}$, define
\begin{equation}
\log p_{\psi}(\mathbf{S})
=
\sum_{j=1}^{n}\log p^{(b)}(\mathbf{s}_{:,j}).
\label{eq:unified_logp_new}
\end{equation}
The evidence lower bound is
\begin{equation}
\mathcal{L}_{\mathrm{ELBO}}
=
\E_{q_{\phi}(\mathbf{S}\mid\mathbf{Y})}[\log p_{\theta}(\mathbf{Y}\mid\mathbf{S})]
-
\KL\big(q_{\phi}(\mathbf{S}\mid\mathbf{Y})\,\|\,p_{\psi}(\mathbf{S})\big).
\end{equation}
In implementation, a single reparameterized sample is used, yielding the normalized objective
\begin{equation}
\mathcal{L}
=
\mathcal{L}_{\mathrm{rec}}
+
\beta
\left[
\log q_{\phi}(\mathbf{S}\mid\mathbf{Y})-\log p_{\psi}(\mathbf{S})
\right],
\label{eq:final_loss_new}
\end{equation}
where $\beta>0$ controls the strength of prior matching \citep{burgess2018understanding,wei2024innovative}.

Equation \eqref{eq:final_loss_new} makes the source-separation mechanism explicit. The reconstruction term alone encourages latent variables that explain the observed mixtures. The prior term alone encourages trajectories that are plausible under source-wise adaptive switching priors. Their joint optimization couples these two pressures. As training proceeds, each latent dimension is repeatedly nudged toward a source trajectory that both reconstructs the mixtures and fits one evolving source-specific prior. Once different dimensions begin to align with different regimes and different prior parameters, the KL term amplifies that divergence rather than collapsing them back together. In this sense, minimizing the loss is simultaneously the process of learning the priors and the process of separating the sources.

\begin{algorithm}[htbp]
\footnotesize
\setlength{\algomargin}{0.8em}
\SetAlgoLined
\DontPrintSemicolon
\SetNlSkip{0.2em}
\SetInd{0.2em}{0.45em}
\caption{Unified training of SAHMM-VAE with branch-specific source scoring}
\label{alg:sahmm_new}
\KwIn{Observed mixtures $\mathbf{Y}=\{\mathbf{y}_t\}_{t=1}^{T}$; number of latent sources $n$; number of hidden states $K$; branch index $b\in\{1,2,3\}$}
\KwOut{Trained $(\phi,\theta,\bm{\sigma}^2,\psi^{(b)})$ and posterior $q_{\phi}(\mathbf{S}\mid\mathbf{Y})$}

Initialize encoder $f_{\phi}$, decoder $g_{\theta}$, posterior variances $\bm{\sigma}^2$, and branch-specific source-prior parameters $\psi^{(b)}=\{\pi_j,A_j,\text{state parameters}\}_{j=1}^{n}$\;

\While{not converged}{
Obtain posterior means $\bm{\mu}_t=f_{\phi}(\mathbf{y}_t)$ for $t=1,\dots,T$\;

Sample latent trajectories using $s_{t,j}=\mu_{t,j}+\sigma_j\epsilon_{t,j}$ with $\epsilon_{t,j}\sim\mathcal{N}(0,1)$\;

Reconstruct observations by $\hat{\mathbf{y}}_t=g_{\theta}(\mathbf{s}_t)$ and compute $\mathcal{L}_{\mathrm{rec}}$ from \eqref{eq:lrec_hmm}\;

Compute posterior score $\log q_{\phi}(\mathbf{S}\mid\mathbf{Y})$ using \eqref{eq:logq_hmm}\;

\For{$j\leftarrow1$ \KwTo $n$}{
Build source-specific hidden-state scorer for trajectory $\mathbf{s}_{:,j}$\;

\eIf{$b=1$}{
Form Gaussian-emission local scores $\ell^{(1)}_{t,j}(k)$ using state means $m_{j,k}$ and variances $v_{j,k}$\;
Run forward recursion $\alpha^{(1)}_{t,j}(k)$ and obtain $\log p^{(1)}(\mathbf{s}_{:,j})=\lse_k\,\alpha^{(1)}_{T,j}(k)$\;
}{
\eIf{$b=2$}{
Form MSAR local scores $\ell^{(2)}_{t,j}(k)$ using $\eta_{t,j,k}=\mu_{j,k}+\phi_{j,k}(s_{t-1,j}-\mu_{j,k})$ and innovation variances $\sigma_{j,k}^2$\;
Run forward recursion $\alpha^{(2)}_{t,j}(k)$ and obtain $\log p^{(2)}(\mathbf{s}_{:,j})=\lse_k\,\alpha^{(2)}_{T,j}(k)$\;
}{
Construct state-flow residuals $u_{t,j,k}=(s_{t,j}-a_{j,k}s_{t-1,j})/\rho_{j,k}$\;
Evaluate inverse-flow noise $\varepsilon_{t,j,k}=f^{-1}_{j,k}(u_{t,j,k})$ and Jacobian term $\log|\partial f^{-1}_{j,k}/\partial u_{t,j,k}|$\;
Form state-flow local scores $\ell^{(3)}_{t,j}(k)$ and run forward recursion $\alpha^{(3)}_{t,j}(k)$\;
Obtain $\log p^{(3)}(\mathbf{s}_{:,j})=\lse_k\,\alpha^{(3)}_{T,j}(k)$\;
}
}
}

Aggregate source-wise prior score $\log p_{\psi}(\mathbf{S})=\sum_{j=1}^{n}\log p^{(b)}(\mathbf{s}_{:,j})$\;

Form final objective
\[
\mathcal{L}=\mathcal{L}_{\mathrm{rec}}+\beta\left[\log q_{\phi}(\mathbf{S}\mid\mathbf{Y})-\log p_{\psi}(\mathbf{S})\right]
\]
Update $(\phi,\theta,\bm{\sigma}^2,\psi^{(b)})$ by gradient descent\;
}

\Return trained parameters and posterior family $q_{\phi}(\mathbf{S}\mid\mathbf{Y})$\;
\end{algorithm}

\begin{figure*}[t]
    \centering
    \includegraphics[width=\textwidth]{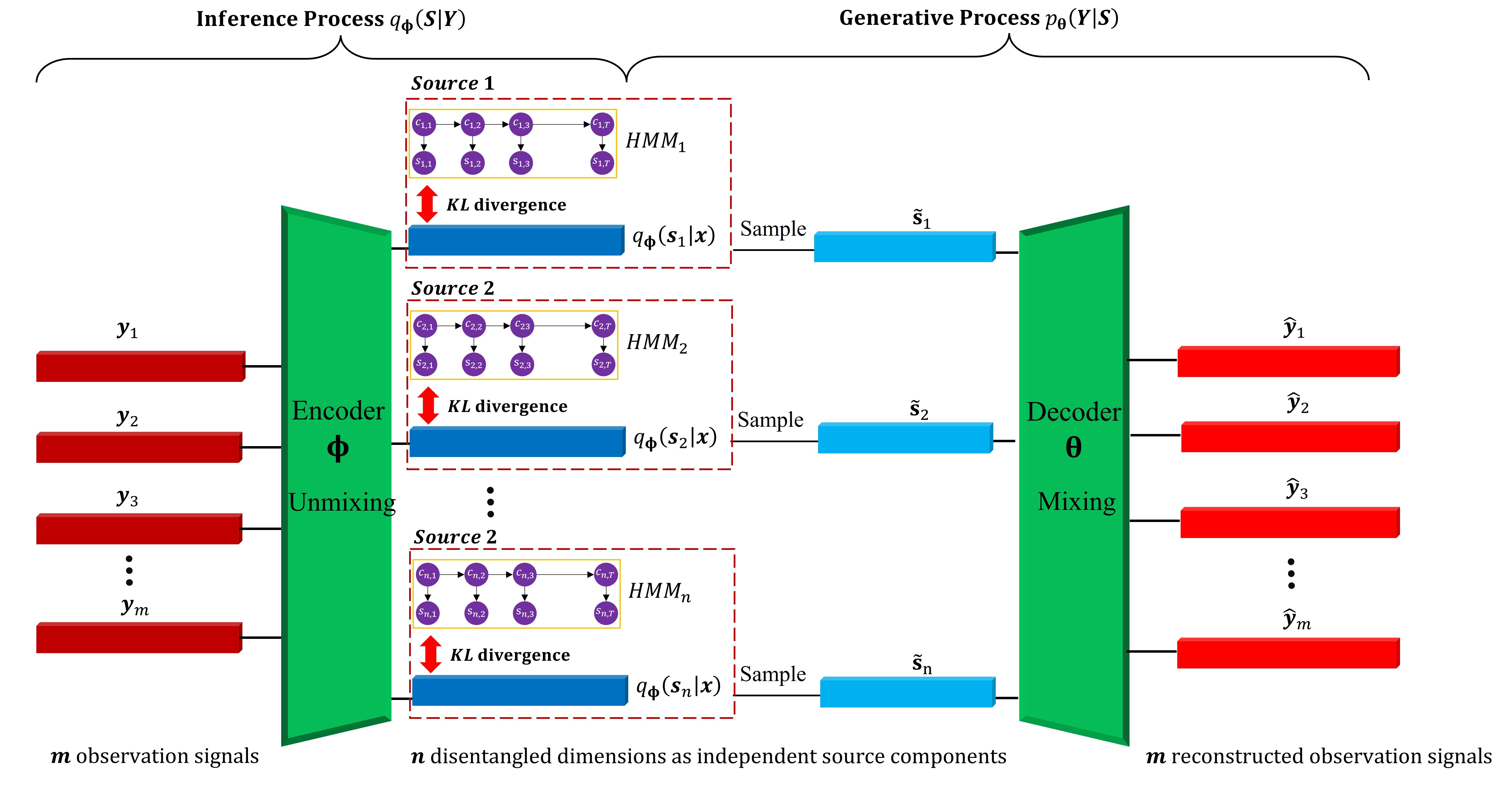}
    \caption{Overall source-wise HMM-prior VAE framework. Each latent dimension is interpreted as one source candidate and is paired with its own adaptive switching HMM prior.}
    \label{fig:framework}
\end{figure*}

\section{Experimental Study}
\label{sec:experiments}

\subsection{Experimental perspective}

The experiments are designed to answer two related questions. The first is whether the proposed source-wise switching-prior design can separate sources accurately under unsupervised training. The second is whether the internal switching structure learned by the priors is meaningfully related to the source dynamics rather than being a purely decorative latent mechanism. For this reason, the results are read from four complementary angles: convergence behavior, source waveform recovery, hidden-state recovery, and consistency between learned transition parameters and empirical transition usage.

Figures~\ref{fig:m1_convergence}--\ref{fig:m3_convergence} report the training traces of the three branches. Figure~\ref{fig:source_recovery} compares the reconstructed latent sources with the ground truth. Figures~\ref{fig:state_sequence} and \ref{fig:source_state_overlay} examine whether the inferred state paths track meaningful regime changes in the sources. Figure~\ref{fig:transition_structure} then compares the learned transition matrices with empirical transition matrices computed from the inferred state sequences. Taken together, these figures allow the quality of separation and the quality of internal prior learning to be discussed jointly rather than in isolation.

\subsection{Method 1: Gaussian-emission HMM prior}
\label{subsec:m1_results_new}

\begin{figure*}[t]
    \centering
    \includegraphics[width=\textwidth]{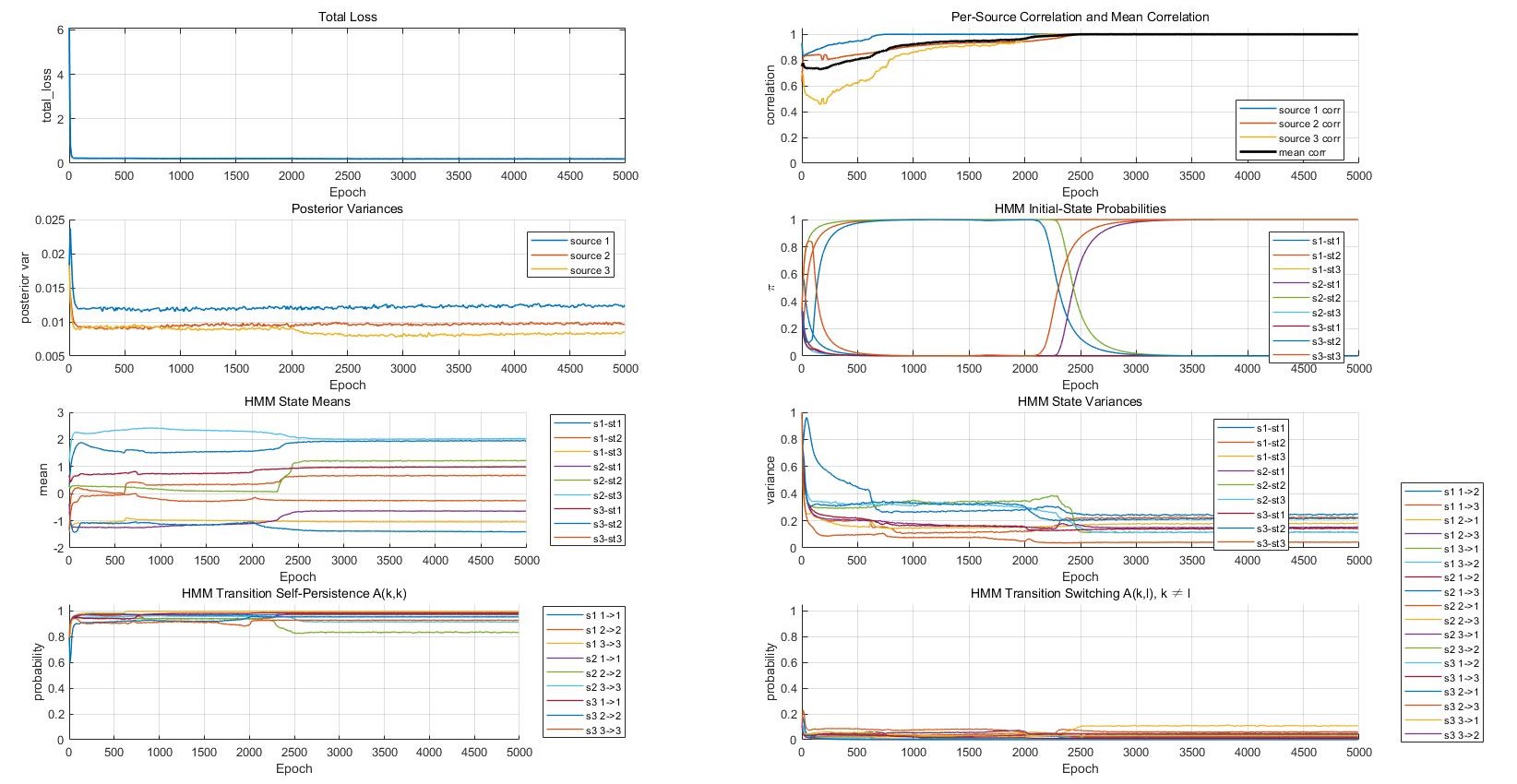}
    \caption{Training convergence of Method 1 (Gaussian-emission HMM prior).}
    \label{fig:m1_convergence}
\end{figure*}

Figure~\ref{fig:m1_convergence} shows that the simplest switching-prior branch already trains stably. The total loss decreases quickly and then settles, while the per-source correlations rise toward values very close to one. This indicates that the model does not need state-dependent autoregression or flow flexibility to begin separating the sources effectively; the source-wise switching bias alone already provides a useful organizing force.

An equally important observation concerns the prior parameters themselves. The state means and state variances move substantially during the earlier stages and then stabilize into separated regimes, while the transition matrices become strongly diagonal-dominant. This is the expected pattern if the model is genuinely learning persistent source-specific regimes rather than merely using the HMM layer as a redundant parametrization. The posterior variances also shrink and stabilize, suggesting that once the decoder and the source-wise priors become mutually compatible, the model requires less posterior uncertainty to explain the observations.

From the standpoint of the paper's main claim, Method 1 already illustrates the central mechanism well. The source estimates improve at the same time as the HMM parameters specialize. In other words, source recovery and prior adaptation are not two disconnected events. The prior is being learned \emph{while} the sources are being separated, and the stabilization of different source-wise HMM parameters is part of the evidence that different latent dimensions are settling into different source roles.

\subsection{Method 2: Markov-switching autoregressive prior}
\label{subsec:m2_results_new}

\begin{figure*}[t]
    \centering
    \includegraphics[width=\textwidth]{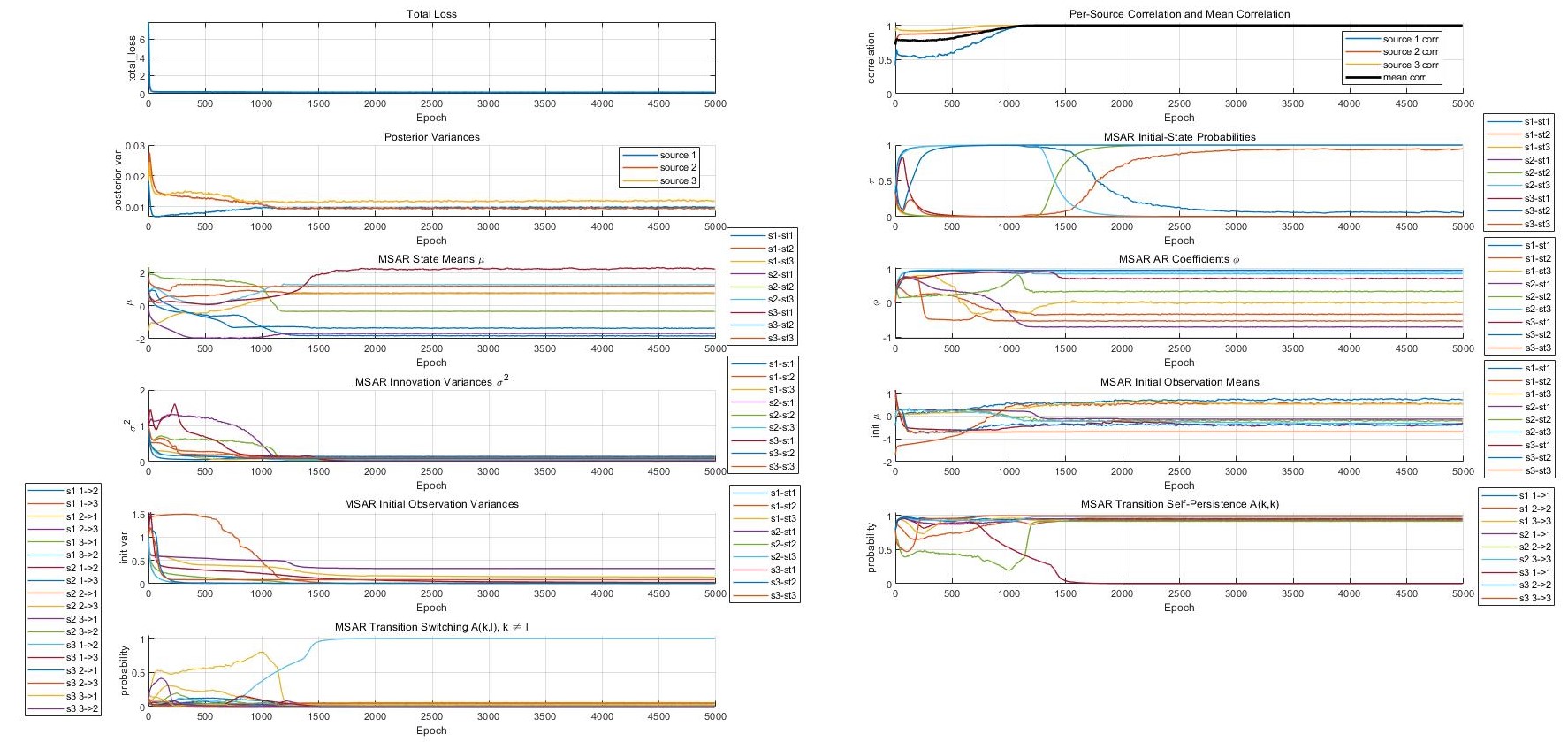}
    \caption{Training convergence of Method 2 (MSAR-HMM prior).}
    \label{fig:m2_convergence}
\end{figure*}

Figure~\ref{fig:m2_convergence} adds a second layer of structure: states are now distinguished not only by marginal level but also by their local temporal law. The optimization remains stable, and the source correlations again approach values extremely close to one. The main difference is that the internal parameter evolution becomes richer. The autoregressive coefficients, innovation scales, and state-dependent means continue adjusting for longer than the parameters of Method 1, which is natural because the model must now decide both \emph{when} states switch and \emph{how} each state propagates local dynamics.

This additional structure matters because many sources are not best distinguished by static amplitude statistics alone. Two regimes may occupy overlapping value ranges and still differ strongly in local persistence or trend. Method 2 is designed for exactly this situation, and the learned AR coefficients settling at clearly different values across states provide evidence that the model is indeed using state-dependent dynamics instead of collapsing back to a purely marginal HMM interpretation.

For the narrative of this paper, Method 2 also sharpens the unsupervised-separation interpretation. The latent dimensions are not merely becoming different because the decoder finds it useful; they are becoming different because each dimension is repeatedly matched to a switching dynamic law whose parameters are also adapting. The separation process is therefore better described as a co-evolution of latent trajectories and source-specific temporal priors than as a plain reconstruction-driven decomposition.

\subsection{Method 3: HMM state-flow prior}
\label{subsec:m3_results_new}

\begin{figure*}[t]
    \centering
    \includegraphics[width=\textwidth]{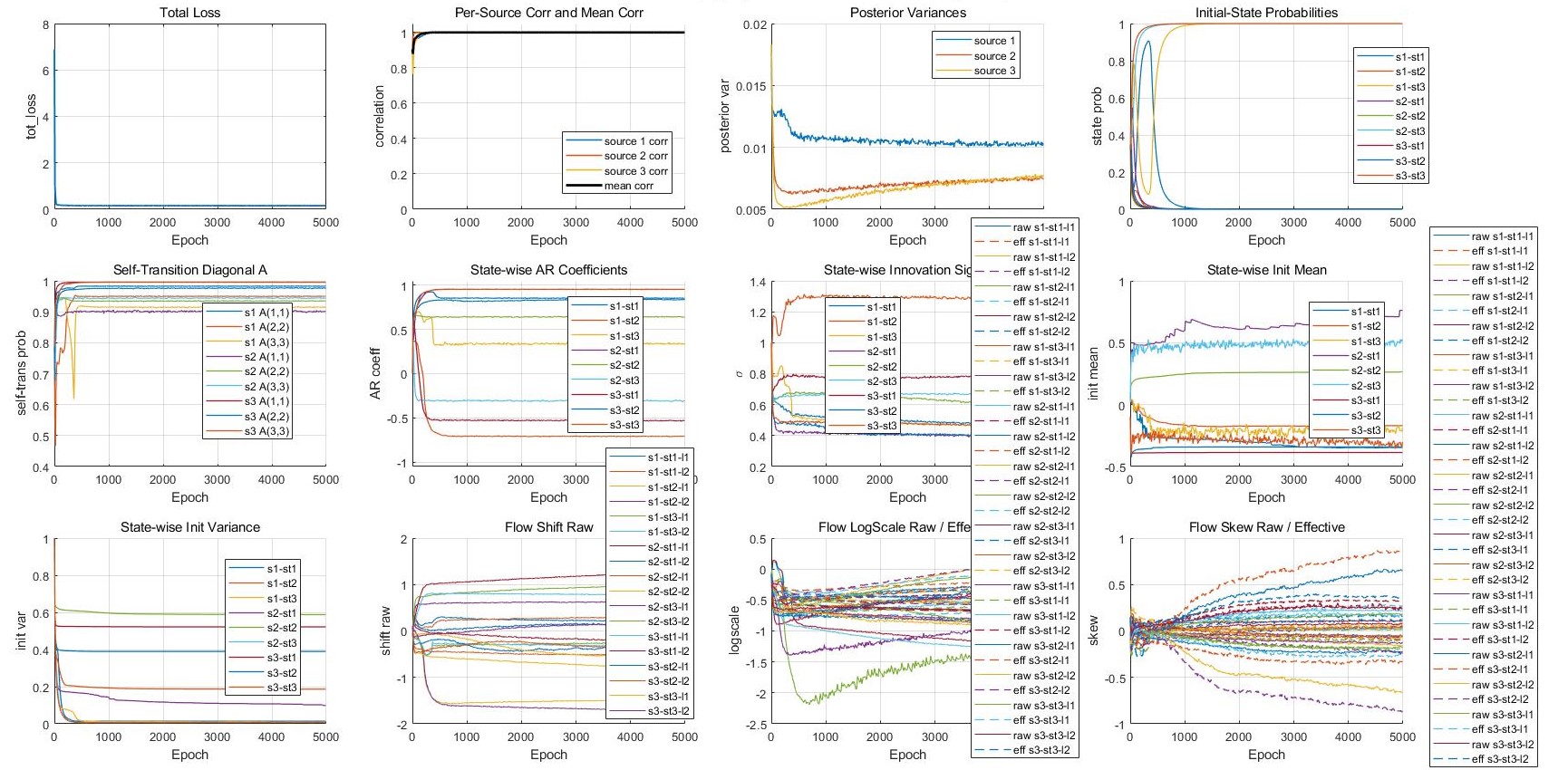}
    \caption{Training convergence of Method 3 (HMM state-flow prior).}
    \label{fig:m3_convergence}
\end{figure*}

Figure~\ref{fig:m3_convergence} corresponds to the most expressive branch. It preserves source-wise Markov switching and state-dependent autoregression, but also equips each state with a flow-based innovation model. Even under this increased flexibility, the total loss still decreases reliably and the source correlations remain extremely high. This confirms that the source-wise switching-prior idea is compatible not only with simple HMMs but also with more expressive state-conditional distributions.

The training traces also reveal a familiar trade-off. Because the branch has greater representational freedom, the internal parameters continue evolving for longer and can admit multiple plausible explanations of the same source waveform. This is especially visible in the state-flow parameters, whose flexibility can absorb local non-Gaussian variation that would otherwise need to be explained by cleaner state partitioning. Consequently, source recovery can remain excellent even when hidden-state interpretability becomes less unique.

This behavior is not a contradiction. It highlights an important distinction running through the paper: accurate source reconstruction and uniquely interpretable latent-state explanation are related but not identical goals. Method 3 leans more heavily toward expressive source modeling. The fact that it can still separate sources well under fully unsupervised training supports the generality of the framework, while the looser state identifiability underscores why switching-prior expressiveness should be discussed together with interpretability rather than only with reconstruction accuracy.

\subsection{Source recovery across all three branches}
\label{subsec:source_recovery_new}

\begin{figure*}[t]
    \centering
    \includegraphics[width=\textwidth]{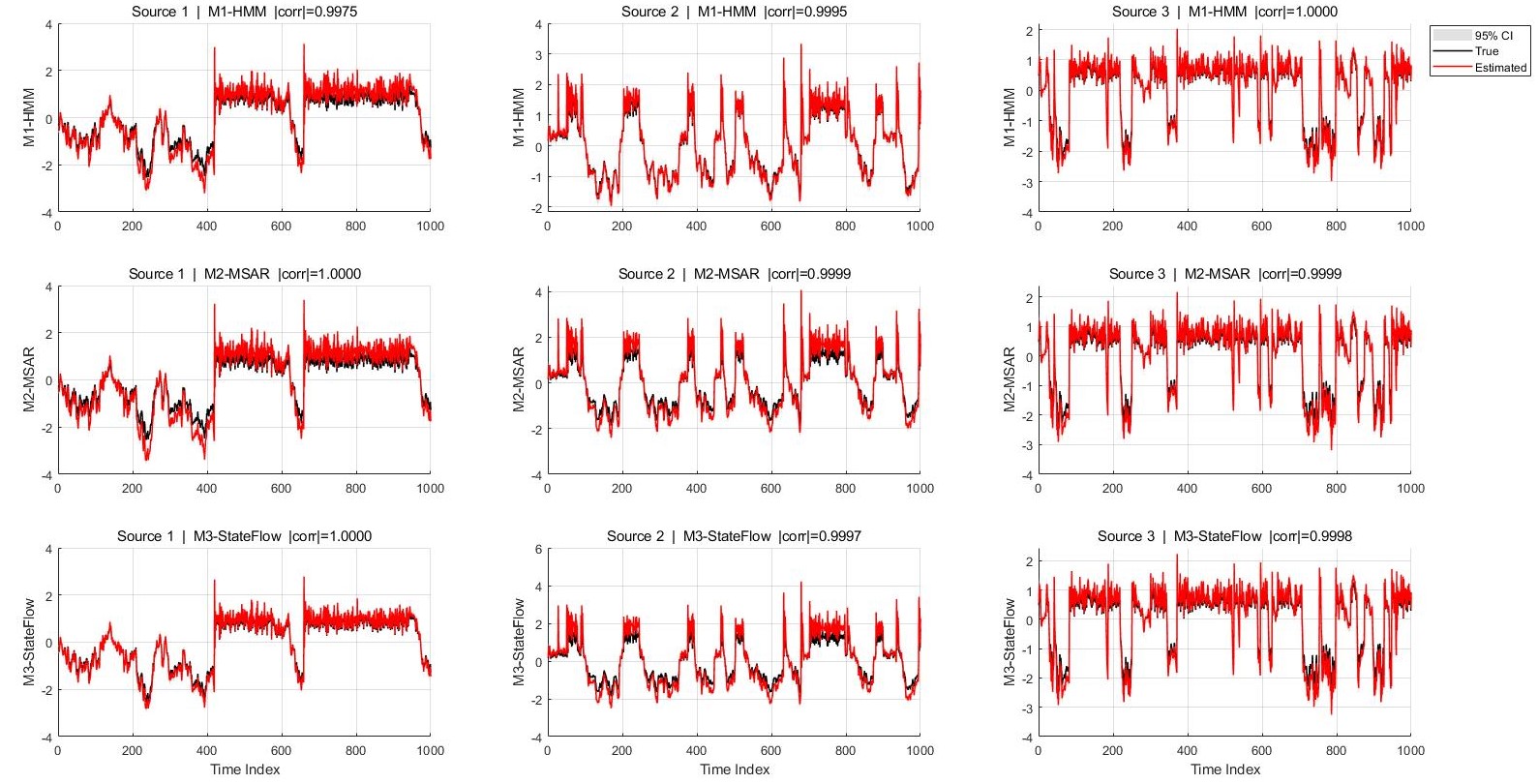}
    \caption{Recovered source signals for the three branches.}
    \label{fig:source_recovery}
\end{figure*}

Figure~\ref{fig:source_recovery} compares the recovered source trajectories from all three branches. At the waveform level, all three methods perform strongly: the estimated trajectories nearly overlap with the true sources and the absolute correlations are all very close to one. This is the most direct evidence that the proposed source-wise HMM-prior VAE family functions effectively as an unsupervised source separator.

The more interesting point is that the visual gap among the methods is smaller than the gap in their internal prior complexity. This means that, for the present data, once each latent dimension is equipped with its own structured switching prior, even the simplest branch already supplies a sufficiently strong inductive bias for accurate separation. The richer branches do not necessarily produce dramatically different waveforms because the baseline itself is already good.

However, that should not be read as evidence that the three priors are interchangeable. Their difference lies in what kind of explanation they impose on the latent trajectories. Method 1 attributes the sources mainly to switching marginal regimes. Method 2 attributes them to switching dynamical regimes. Method 3 allows more flexible non-Gaussian state-wise innovation structure. Thus, Figure~\ref{fig:source_recovery} is best interpreted as validating the common separation principle, while the later figures explain how the three branches differ in internal temporal interpretation.

\subsection{Hidden-state recovery}
\label{subsec:state_recovery_new}

\begin{figure*}[t]
    \centering
    \includegraphics[width=\textwidth]{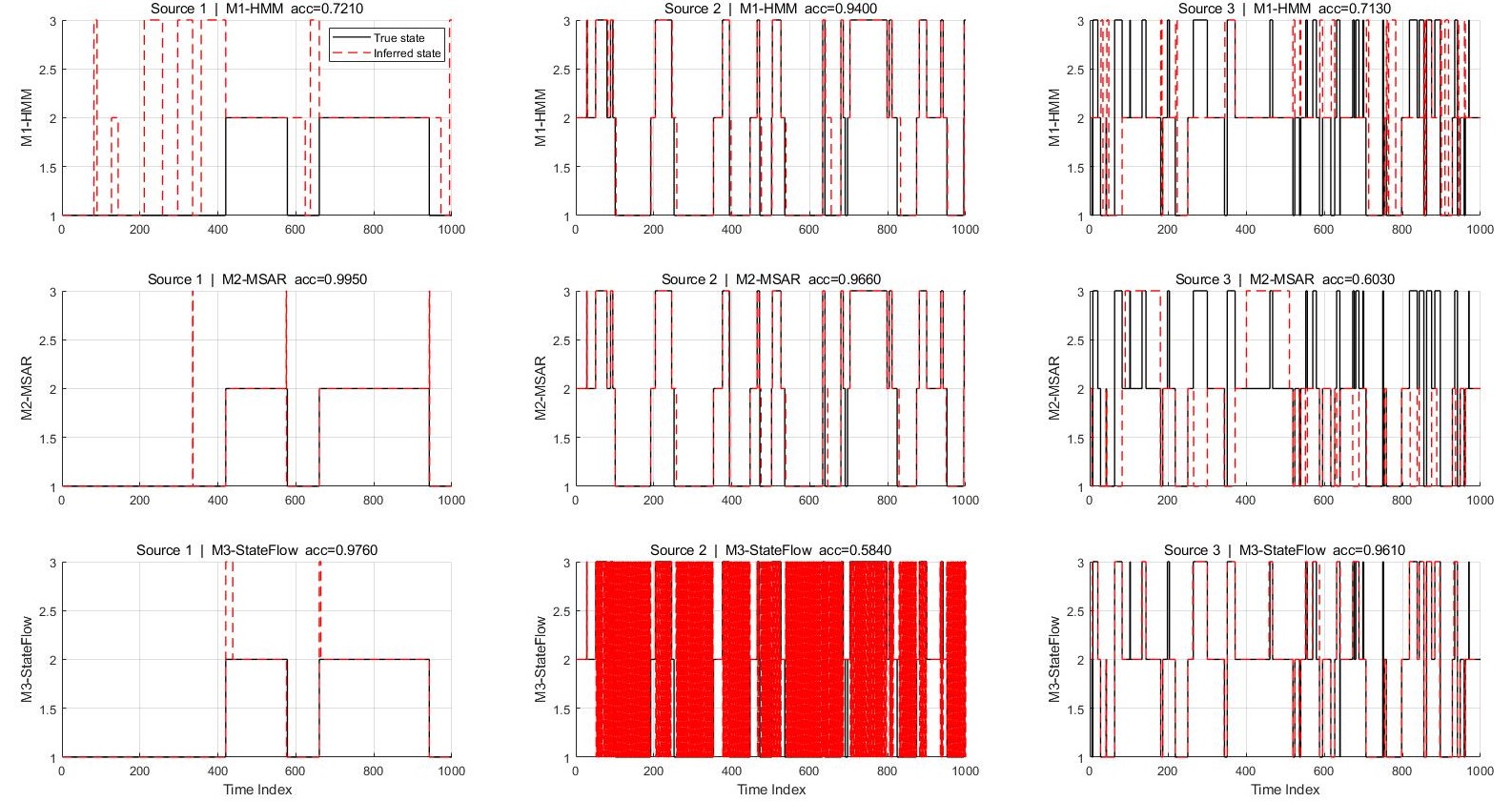}
    \caption{Comparison between true and inferred hidden-state sequences.}
    \label{fig:state_sequence}
\end{figure*}

Figure~\ref{fig:state_sequence} compares true and inferred hidden-state paths. This result is especially informative because it moves beyond source waveform recovery and asks whether the learned priors are capturing the intended regime structure.

For Method 1, state recovery is reasonable but uneven across sources. This is expected because a Gaussian-emission HMM distinguishes states primarily through state-dependent marginal distributions. When two regimes occupy overlapping value ranges or differ more in local dynamics than in static amplitude, state discrimination becomes harder. Even so, the inferred paths still often track the broad switching pattern of the true sources, which indicates that the source-wise HMM layer is not arbitrary.

Method 2 generally improves the picture when the true source regimes differ by local linear dynamics. In such cases, state-dependent autoregression provides an extra cue that helps distinguish states that would otherwise look similar in amplitude alone. This is precisely where the added structure of Method 2 becomes most meaningful: it uses temporal evolution, not only instantaneous value, to decide which regime is active.

Method 3 presents a more mixed result. Some sources can exhibit near-perfect waveform recovery while their inferred state sequence is less well aligned with the ground-truth state path. This is not surprising under a flexible state-flow prior. The model can distribute explanatory burden between switching structure and state-wise nonlinear transformations, so a unique discrete interpretation is no longer always necessary for high-quality source reconstruction.

Another practical issue is HMM state-label ambiguity. Hidden states are identifiable only up to permutation, and different parameterizations may support very similar likelihoods. Therefore, imperfect state-sequence matching should not automatically be interpreted as model failure. Rather, it reflects the broader lesson that source recovery quality and state identifiability are distinct but interacting properties.

\subsection{Local source--state overlay}
\label{subsec:overlay_new}

\begin{figure*}[t]
    \centering
    \includegraphics[width=\textwidth]{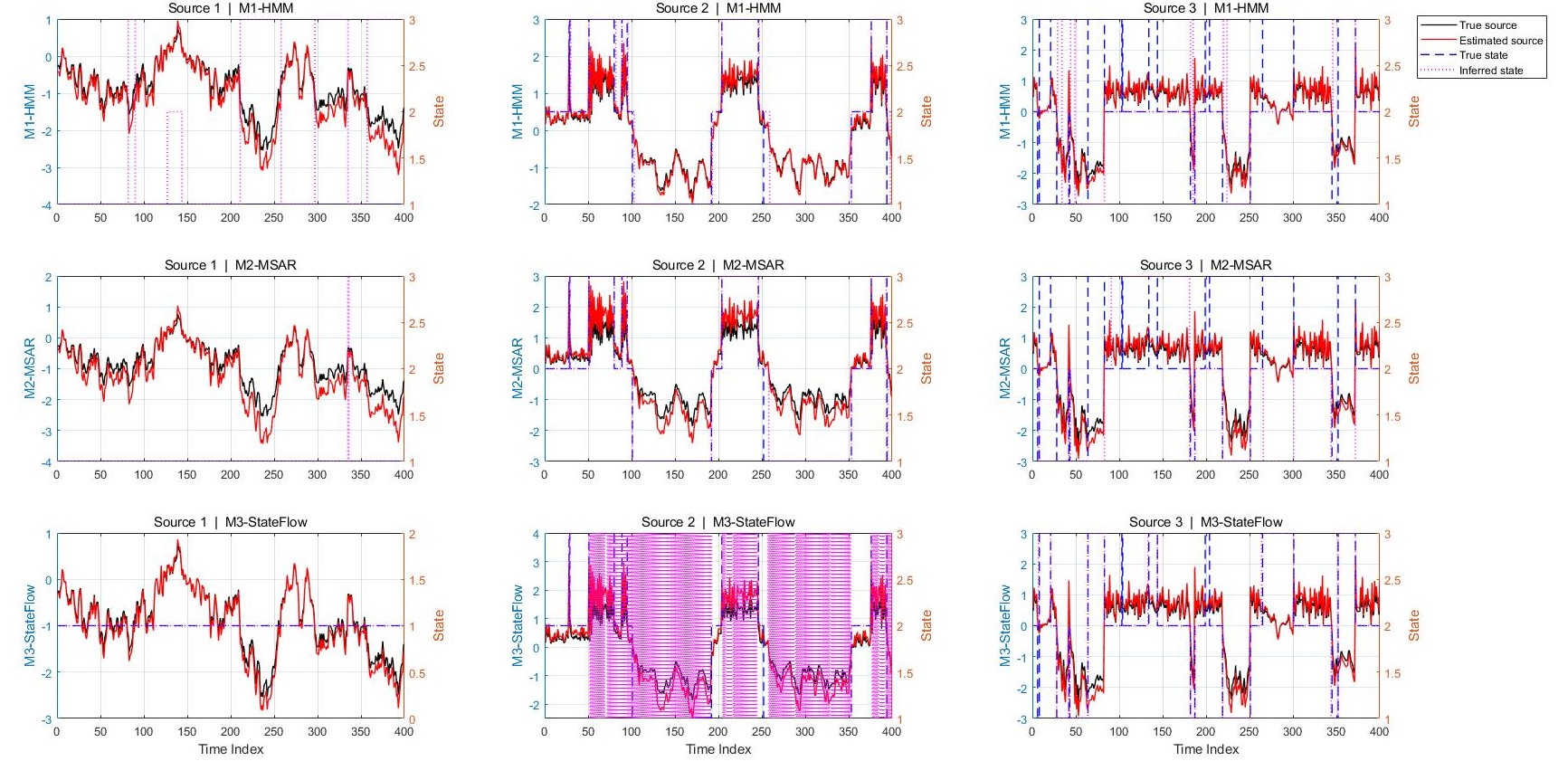}
    \caption{Local overlay of source reconstruction and hidden-state paths.}
    \label{fig:source_state_overlay}
\end{figure*}

Figure~\ref{fig:source_state_overlay} offers a local view by plotting source trajectories together with their associated true and inferred state paths. This figure is useful because it shows \emph{where} the inferred switching structure is being used.

For Methods 1 and 2, many inferred state changes occur near visually meaningful changes in signal level or local temporal pattern. This supports the interpretation that the hidden states are responding to actual source structure rather than to arbitrary numerical fluctuations. Method 2 is especially interpretable in segments where local persistence changes, because the inferred states often align with switches in the local temporal law itself.

For Method 3, the source fit remains strong, but the inferred state path can become denser or less visually synchronized with the true state partition. This again reflects the extra freedom of the state-flow branch. The model can explain local irregularities through the flow transformation rather than by reserving discrete states only for clean regime boundaries. Hence, the source path can be accurate even when the discrete state path is less transparently matched.

The figure therefore reinforces one of the main conceptual messages of the paper: simpler switching priors may provide cleaner hidden-state explanations, whereas more expressive priors may trade some of that clarity for greater modeling flexibility, even when all of them remain effective separators.

\subsection{Learned versus empirical transition structure}
\label{subsec:transition_results_new}

\begin{figure*}[t]
    \centering
    \includegraphics[width=\textwidth]{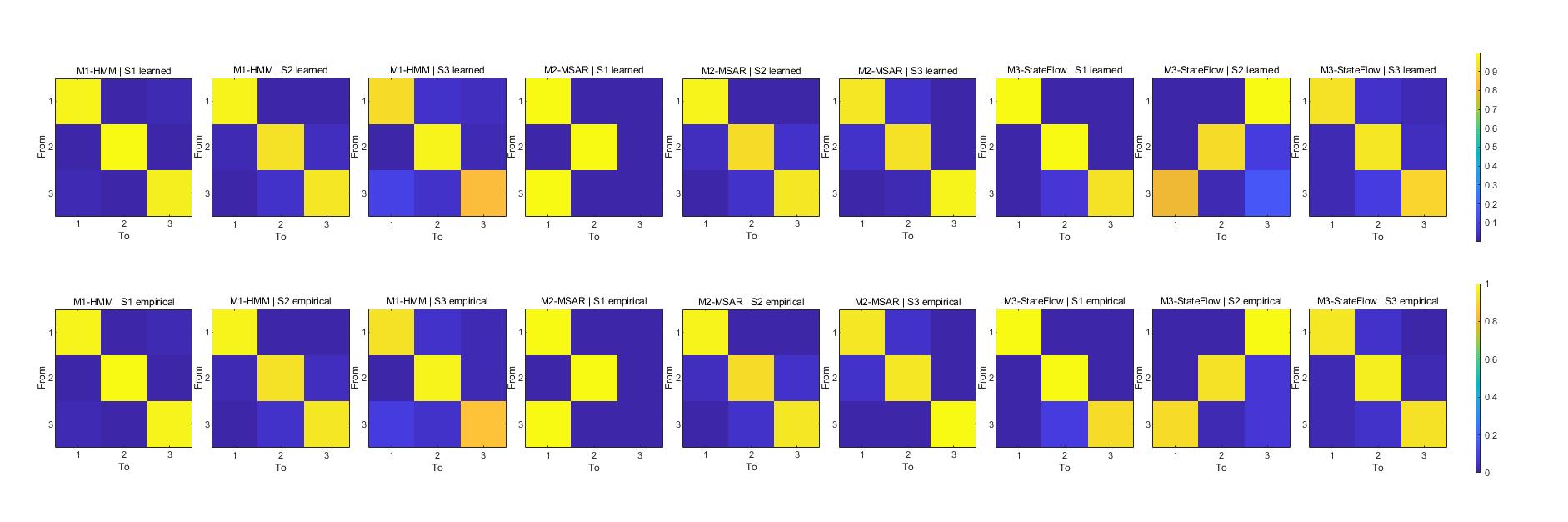}
    \caption{Learned and empirical transition matrices for all methods and sources.}
    \label{fig:transition_structure}
\end{figure*}

Figure~\ref{fig:transition_structure} compares the learned transition matrices with empirical transition matrices computed from the inferred state sequences. The purpose is not to demand exact equality, but to check whether the learned source-wise priors are internally consistent with how the model actually uses its states.

In most cases, the qualitative agreement is strong. Both the learned and empirical matrices are typically diagonal-dominant, indicating persistent regimes with relatively sparse switching. This is exactly the kind of source-wise regime organization the model is supposed to learn. The agreement therefore supports the view that the adaptive HMM priors are not merely being optimized numerically; they are shaping actual state usage in the inferred latent trajectories.

Discrepancies are nonetheless expected. The learned transition matrix is a parameter of the prior model, whereas the empirical matrix is computed from one inferred hard state path, often after Viterbi-type decoding or posterior mode assignment. These two objects are related but not identical. They can differ because of posterior uncertainty, finite-sample variation, state-label permutation, and the fact that different hard state paths may be compatible with similar soft posterior structure.

This distinction becomes more pronounced in expressive branches such as Method 3. When the prior can represent a source through several internally plausible state-flow combinations, one inferred state path may not reproduce the nominal transition parameters numerically even if both convey the same broad switching logic. For this reason, the figure is most informative when read qualitatively. Its main message is that the learned source-wise priors and the realized state usage largely point to the same type of regime structure.

\subsection{Overall discussion}

Taken together, the experiments support three conclusions. First, all three branches converge stably and recover the source waveforms accurately. This validates the main modeling premise of the paper: assigning each latent dimension its own adaptive switching prior is an effective unsupervised route to source separation. Second, the three branches differ mainly in \emph{how} they explain the sources internally rather than in whether they can separate them at all. Third, increasing prior expressiveness does not automatically improve hidden-state interpretability; in some cases it weakens the uniqueness of the discrete explanation while preserving or even improving source recovery. This is an important observation for future work on identifiability and interpretable latent dynamics.

\section{Conclusion and Future Work}
\label{sec:conclusion}

This paper proposed SAHMM-VAE, a source-wise adaptive HMM-prior variational autoencoder for unsupervised blind source separation. The defining idea is to treat each latent dimension as an individual source candidate and to couple it with its own adaptive switching prior. Under this design, the latent prior is no longer a passive regularizer shared across all dimensions. Instead, it becomes a source-specific structural force that co-evolves with the posterior trajectories and decoder during training.

Within this common framework, we studied three progressively richer branches: a Gaussian-emission HMM prior, a Markov-switching autoregressive HMM prior, and an HMM state-flow prior. Across all three branches, experiments showed that optimizing the variational objective yields accurate source recovery and stable training. Just as importantly, the learned source-wise priors and hidden-state structures provide evidence that source separation is occurring together with prior adaptation rather than after it. As different latent dimensions settle into different prior parameters and different temporal organizations, the KL coupling between posterior and prior becomes part of the separation mechanism itself.

The comparison among the three branches also clarifies a broader modeling trade-off. Simpler switching priors may offer cleaner hidden-state interpretation, whereas more expressive state-wise priors can better absorb complex source variability but may reduce the uniqueness of the inferred regime explanation. This suggests that future structured-prior source models should be assessed not only by reconstruction quality or separation accuracy, but also by the transparency and identifiability of the latent temporal structure they induce.

Several directions follow naturally from the present work. One is to study the identifiability of source-wise adaptive switching priors more formally, especially when different latent dimensions are governed by heterogeneous temporal laws. Another is to develop stronger constraints or regularizers that preserve the flexibility of expressive branches while improving state interpretability. It is also worthwhile to test the framework under more challenging nonlinear, noisy, or underdetermined settings. Finally, a deeper theoretical analysis of convergence, prior--posterior co-adaptation, and the conditions under which source-wise prior differentiation guarantees more robust separation would further strengthen the foundation of this research line.

\clearpage
\bibliography{ref}

\end{document}